\documentclass{article} 
\usepackage{iclr2027_conference,times}

\usepackage[utf8]{inputenc} 
\usepackage[T1]{fontenc}    
\usepackage[dvipsnames,table]{xcolor}
\usepackage{hyperref}       
\usepackage{url}            
\usepackage{booktabs}       
\usepackage{amsmath}
\usepackage{amsfonts}       
\usepackage{nicefrac}       
\usepackage{microtype}      
\usepackage{multirow}
\usepackage{graphicx}
\usepackage{pifont}
\usepackage{makecell}
\definecolor{mycolor}{RGB}{220,230,241} 
\definecolor{Wu}{RGB}{0,0,255}

\title{Entropogram-Guided Visual Realignment Decoding: Capturing the Rhythm of Multimodal Reasoning}

\author{Jiaqi Deng$^{1*}$, Zhe Fu$^{1*}$, Yakun Chen$^{2\dagger\ddagger}$, Dichang Zhang$^{3}$, Zonghan Wu$^{4\ddagger}$, \\
\bf Xiaoshui Huang$^{5}$, Huan Huo$^{1}$, Guandong Xu$^{1,6}$ \\
\normalfont $^{1}$University of Technology Sydney \quad $^{2}$Alibaba Cloud Computing \quad $^{3}$Stony Brook University \\
\normalfont $^{4}$East China Normal University \quad $^{5}$Shanghai Jiao Tong University \\
\normalfont $^{6}$The Education University of Hong Kong
}

\iclrpreprint 
\begin{document}

\maketitle
{\renewcommand{\thefootnote}{$*$}\footnotetext{Equal contribution (co-first authors).}}
{\renewcommand{\thefootnote}{$\dagger$}\footnotetext{Work done while at The Education University of Hong Kong.}}
{\renewcommand{\thefootnote}{$\ddagger$}\footnotetext{Corresponding authors.}}

\begin{abstract}
Multimodal large language models (MLLMs) have demonstrated remarkable progress on cross-modal reasoning tasks, yet they continue to suffer from hallucinations in long-chain reasoning processes. In this work, we revisit hallucination in MLLMs from the perspective of reasoning generation dynamics. We argue that entropy is not merely a static measure of uncertainty but a structured signal that reflects how modality information is integrated throughout decoding. Through systematic analysis, we identify two characteristic entropy patterns associated with hallucination behaviors: (\textbf{1}) a generally high-confidence regime corresponds to premature convergence to inappropriate multimodal interactions; and (\textbf{2}) local abrupt entropy transitions are associated with reasoning instability and hallucination onset. Building on these observations, we propose an entropy-aware self-introspective framework for MLLM reasoning. Our method continuously tracks entropy dynamics during decoding and performs targeted visual re-alignment when suspicious hallucination states are detected, encouraging the model to re-examine salient visual evidence at critical reasoning stages. Extensive experiments across six multimodal reasoning benchmarks spanning nine distinct evaluation settings demonstrate that our approach delivers robust improvements in reasoning accuracy and hallucination robustness, while providing new insights into the generation dynamics underlying MLLM failures. The code is available at: \url{https://github.com/jden4524/SelfRetrospectiveDecoding}
\end{abstract}

\section{Introduction}
Multimodal large language models (MLLMs) have achieved significant progress on cross-modal reasoning tasks~\citep{Wu2024GPT-4VisionGeneration, Deitke2024MolmoModels, Zhu2025InternVL3:Models,Bai2025Qwen3-VLReport}, yet they continue to suffer from hallucinated outputs that are unfaithful to the provided visual context~\citep{Li2023EvaluatingModels,Leng2024MitigatingDecoding, Huang2024OPERA:Retrospection-Allocation, Liu2024AModels}. The issue becomes particularly pronounced in long-horizon reasoning processes such as Chain-of-Thought (CoT), where errors at intermediate steps can propagate and compound.

Existing efforts largely attribute such failures to insufficient reasoning or weak representations, and therefore attempt to mitigate them through stronger supervision signals such as visual-reward reinforcement learning~\citep{Liu2024MitigatingTuning, Sun2025LatentReasoning, Pan2026GroundRegularization,Deng2026Mind}. While effective, these approaches necessitate expensive training. Consequently, recent studies also explore training-free interventions, including contrastive decoding and reasoning-aware decoding strategies~\citep{Leng2024MitigatingDecoding, Xu2025MitigatingDecoding, Xu2026ThinkingDecoding}, which improve generation by identifying hallucination-associated tokens and applying interventions to suppress their impact.
Despite their effectiveness, these methods predominantly address \emph{what} tokens are hallucinated, but do not investigate \emph{how} multimodal interactions shape and steer decoding trajectories, which can ultimately lead to hallucinations.

\begin{figure*}[t]
    \centering
    \includegraphics[width=\textwidth]{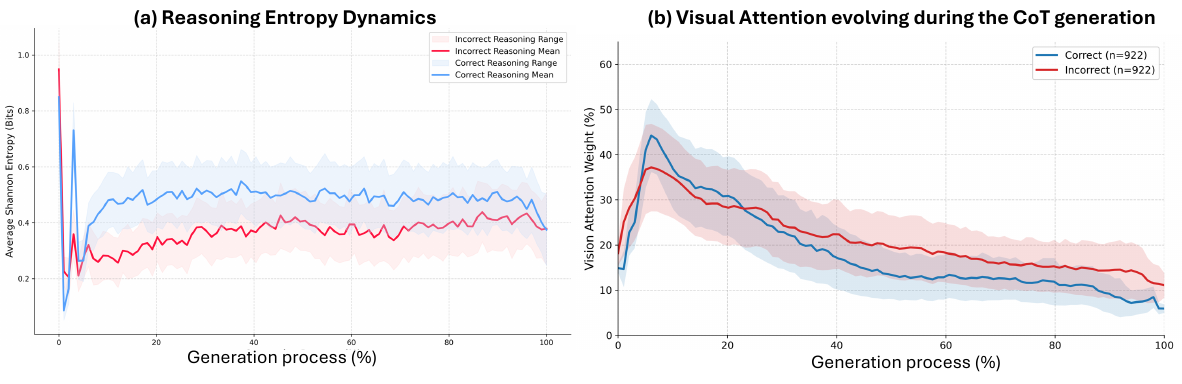}
    \caption{Comparison of generation dynamics between correct and incorrect multimodal reasoning trajectories. (a) Average reasoning entropy across the generation process over samples, where correct predictions consistently maintain higher entropy than incorrect ones. (b) Evolution of visual attention during CoT generation, showing that correct reasoning trajectories preserve stronger visual grounding at the early stage of reasoning, while incorrect trajectories exhibit progressively weakened but still higher attention to visual evidence. Shaded regions denote standard deviations.}
    \label{fig:preliminary1}
\end{figure*}

\begin{figure*}[t]
    \centering
    \includegraphics[width=\textwidth]{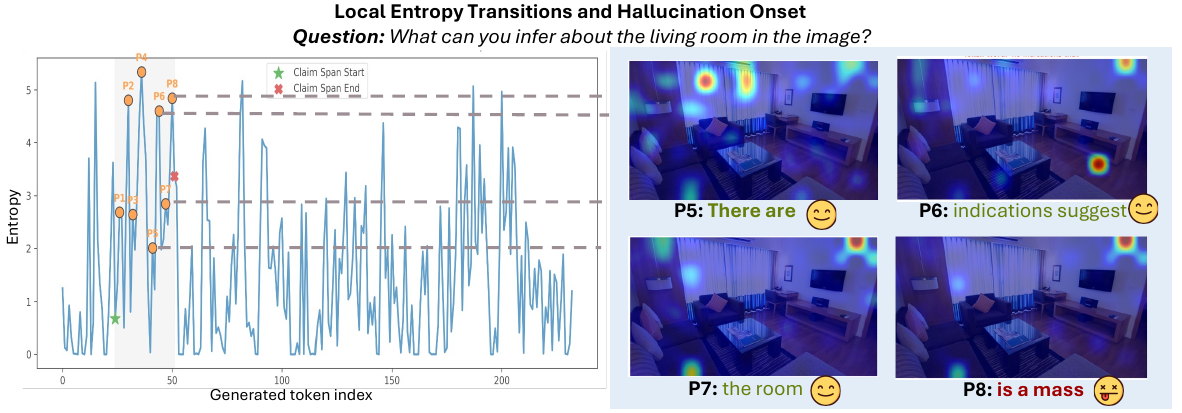}
    \caption{A case study illustrating hallucination emergence during MLLM reasoning. The left panel shows entropy over generated tokens, while the right panel visualizes the corresponding visual attention patterns at selected reasoning steps (P5--P8). The gray region denotes the span of incorrect generation. Hallucinated reasoning coincides with sharp entropy spikes (P8) and progressively strong but diffused visual attention. More examples can be found in Appendix~\ref{preliminary}.}
    \label{fig:preliminary2}
\end{figure*}

In this work, we revisit MLLM hallucination from the perspective of entropy-driven generation dynamics. While recent studies have begun leveraging entropy for decoding in both unimodal~\citep{He2026ThinkReasoning, Sharma2025ThinkReasoning, Wang2025BeyondReasoning, Cheng2026ReasoningPerspective,Zhu2026EDIS:Dynamics} and multimodal~\citep{Xu2026ThinkingDecoding,Zou2025LookModels} settings, they primarily treat entropy as a token-level confidence signal. We instead argue that entropy encodes richer, trajectory-level information about how modality information is integrated over time. Through systematic analysis, we identify two recurring entropy patterns that statistically distinguish correct from hallucinated generation: (1)~prolonged low-entropy (high-confidence) regimes, which counterintuitively correspond to ineffective multimodal consolidation and premature trajectories; and (2)~abrupt entropy transitions associated with local instability and hallucination onset (see Figures~\ref{fig:preliminary1}--\ref{fig:preliminary2} and Table~\ref{tab:population}). These findings indicate that hallucination is closely associated with dynamic failures in cross-modal interaction over time, beyond static reasoning difficulty alone.

Building on this insight, we propose \textbf{SEED} (\textbf{S}elf-introspective \textbf{E}ntropy-aware visual r\textbf{E}-alignment \textbf{D}ecoding), an entropy-aware framework for MLLM reasoning. SEED continuously monitors the entropy dynamics of the reasoning trajectory during decoding; when the entropy signal indicates that generation is entering a hallucination-prone regime, SEED performs targeted re-alignment of visual attention, effectively correcting over-reliance on language priors at critical moments. Unlike prior approaches that intervene uniformly or rely on fixed heuristics, SEED adapts its intervention to the evolving state of cross-modal grounding. In summary, our contributions are threefold:
\begin{itemize}
    \item We provide a systematic empirical analysis showing that hallucinations in MLLMs are consistently associated, at the population level, with two characteristic entropy patterns: sustained low-entropy convergence and abrupt entropy transitions, indicating an association with disrupted modality interaction dynamics beyond token-level uncertainty alone.
    \item We propose SEED, a training-free, entropy-aware decoding framework that monitors generation dynamics in real time and performs targeted visual re-alignment when hallucination-prone states are detected, directly addressing the identified failure modes.
    \item We conduct extensive experiments across multiple base MLLMs and diverse benchmarks. Results demonstrate that SEED delivers broad and substantial improvements in reasoning accuracy and hallucination robustness over strong baselines, introducing less than 5\% computational overhead.
\end{itemize}


\section{The Pattern of Entropy Dynamics in MLLM Reasoning}
To understand the intrinsic mechanism underlying MLLM reasoning, we use information entropy to characterize the decoding process. While entropy is commonly interpreted as a measure of prediction uncertainty or LLMs' confidence, we extend this perspective by asking whether entropy trajectories can reveal how modalities interact during generation. Specifically, we employ Qwen3-VL-8B-Instruct~\citep{Bai2025Qwen3-VLReport} to conduct CoT reasoning on the M$^3$CoT~\citep{Chen2024M3CoT:Chain-of-Thought} benchmark. During reasoning, the entropy at each decoding step $t$ is tracked and calculated as $H_t = -\sum_{w\in\mathcal{V}} p(w\mid x_{<t})\log p(w\mid x_{<t})$, where $x_{<t}$ denotes the preceding token sequence and $\mathcal{V}$ the vocabulary.

\paragraph{Higher controlled uncertainty supports more effective MLLM reasoning.} The first observation arises from 2,000 randomly selected Visual Question Answering (VQA) instances, stratified into correctly and incorrectly solved ones. As shown in Figure~\ref{fig:preliminary1}(a), we observe a consistent entropy-level separation: correctly answered samples maintain persistently higher entropy throughout the reasoning trajectory compared to incorrect ones. Interestingly, this trend remains stable across all decoding stages, suggesting that successful reasoning is not associated with maximal confidence, but rather with sustained uncertainty during generation.

We further analyze the evolution of visual attention during reasoning. Figure~\ref{fig:preliminary1}(b) shows that correctly answered samples initially allocate more attention to visual tokens, which then quickly decays and stabilizes at a relatively low level. In contrast, incorrect cases exhibit a slower decay of visual attention, maintaining higher visual attention in later stages of generation. These two findings are temporally coupled and suggest a strong connection between entropy dynamics and modality interaction: higher-entropy trajectories correspond to faster convergence of visual attention, while lower-entropy trajectories correspond to prolonged but ineffective visual dependency. This alignment suggests that entropy not only reflects uncertainty in reasoning generation, but also implicitly captures how rapidly and effectively visual information is integrated into the reasoning process.

\paragraph{Localized entropy spikes mark hallucination onset.}
To contextualize how hallucinations emerge during generation, we first conduct an instance-level trajectory analysis. One example is illustrated in Figure~\ref{fig:preliminary2}, where the ``\texttt{tidy room}'' has been falsely interpreted as ``\texttt{the room is a mess}''. The hallucination emerges progressively over the course of the reasoning trajectory. Prior to the exact hallucinated content, the entropy curve exhibits multiple sharp spikes (P2, P4, P6, and P8), which correspond to the largest-magnitude transitions observed across the entire decoding process. Meanwhile, the model gradually loses concentration on relevant visual evidence and begins generating unsupported claims. Notably, the hallucinated reasoning span coincides with the unstable entropy transitions and the attention diffusion, suggesting a localized breakdown of multimodal coordination during reasoning.

\paragraph{Population-level evidence.} To verify that this pattern extends beyond individual cases, we analyze 1,228 randomly sampled VQA trajectories, including 763 hallucinated responses and 465 correct controls. For each hallucinated response, we align the first hallucinated token to position $p{=}0$ and compute the token-to-token entropy change $\Delta H_t=H_t-H_{t-1}$ within a $\pm20$-token window; for each correct response, $p{=}0$ is sampled uniformly at random. At hallucination onset, the mean entropy jump is more than $20\times$ larger than in correct trajectories, while differences before the onset are substantially smaller (Table~\ref{tab:population}; full profile in Appendix~\ref{app:population}). Across the complete trajectories, we further measure the longest low-entropy run, the fraction of steps with $\Delta H_t>\tau_{\text{rise}}$, and the maximum entropy change. Hallucinated responses show higher values on all three measures, with significantly longer low-entropy runs, more frequent spikes, and larger maximum changes (two-sided Mann--Whitney U tests, all $p<10^{-3}$). These consistent associations motivate SEED’s use of entropy dynamics as a proxy signal for hallucination-prone states.

\begin{table}[t]
\centering
\small
\caption{Population-level entropy statistics on 1,228 VQA trajectories (Qwen3-VL-8B). $\Delta H$@onset is the mean entropy change at the aligned hallucination onset ($p{=}0$); the remaining columns are per-trajectory statistics. $p$-values are from two-sided Mann--Whitney U tests.}
\label{tab:population}
\setlength{\tabcolsep}{5pt}
\begin{tabular}{lcccc}
\toprule
 & \textbf{$\Delta H$@onset} & \textbf{Low-entropy duration} & \textbf{Spike frequency} & \textbf{Max $\Delta H$} \\
\midrule
Correct ($n{=}465$) & 0.012 & 17.6 & 0.085 & 1.762 \\
Hallucinated ($n{=}763$) & 0.245 & 21.4 & 0.101 & 1.974 \\
\midrule
$p$-value & $4\times10^{-4}$ & $<10^{-35}$ & $<10^{-9}$ & $<10^{-16}$ \\
\bottomrule
\end{tabular}
\end{table}

\ding{46}~To summarize, the two findings suggest that entropy trajectories encode substantially richer information than prediction uncertainty alone. Beyond serving as a confidence signal, entropy appears to implicitly reflect the evolving state of multimodal information consolidation throughout reasoning. Existing entropy-based decoding approaches largely inherit assumptions from unimodal LLMs and therefore interpret entropy primarily as a token-level uncertainty indicator. Our observations instead suggest that, in MLLMs, entropy also captures the stability of cross-modal grounding dynamics. We therefore hypothesize that \textit{effective multimodal reasoning requires maintaining a controlled entropy regime that enables stable multimodal information consolidation throughout generation.}

\section{Literature Review}
\subsection{Multimodal Large Language Models}
Recent MLLMs have made substantial progress in multimodal reasoning, which requires both integrating visual perception with language understanding and reasoning over multimodal clues~\citep{Wu2024GPT-4VisionGeneration, Deitke2024MolmoModels, Zhu2025InternVL3:Models,Bai2025Qwen3-VLReport}. A typical MLLM architecture consists of a modality alignment module (e.g., Q-Former~\citep{Li2023BLIP-2:Models}) bridging visual encoders (e.g., ViT-L/14~\citep{Dosovitskiy2020AnScale}) and LLMs (e.g., Flan-T5~\citep{Raffel2019ExploringTransformer}, GPT-3~\citep{Brown2020LanguageLearners}).

To further enhance reasoning capabilities, Chain-of-Thought (CoT) prompting~\citep{Zheng2023DDCoT:Models,Gao2025Interleaved-ModalChain-of-Thought, Zhang2024MultimodalModels} and supervised fine-tuning (SFT)~\citep{Dai2023InstructBLIP:Tuning, Liu2023VisualTuning, Bai2025Qwen3-VLReport} have been widely adopted to refine step-by-step reasoning behaviors. More recently, reinforcement learning, such as Group Relative Policy Optimization (GRPO), and process-level supervision have been introduced to improve reasoning faithfulness and long-horizon planning abilities~\citep{Zhou2025ReinforcedModels, DeepSeek-AI2026DeepSeek-R1:Learning,Deng2025ATask, Song2025RLSFT, Xing2026CAPRL:LEARNING}.

\subsection{Mitigating Hallucinations in MLLM Reasoning}
Hallucinations in MLLMs refer to outputs that are either inconsistent with the multimodal context or entirely ungrounded, including incorrect object attribution, fabricated attributes, and irrelevant reasoning~\citep{Li2023EvaluatingModels, Liu2024AModels}. Unlike hallucinations in language models, multimodal hallucinations arise from the intricate coupling between visual perception and language generation, making their causal factors harder to disentangle and mitigate.

In the context of large language models, a substantial body of work has studied hallucination mitigation through inference-time decoding strategies~\citep{He2026ThinkReasoning, Sharma2025ThinkReasoning, Wang2025BeyondReasoning, Cheng2026ReasoningPerspective,Zhu2026EDIS:Dynamics}. These approaches leverage internal confidence signals, such as entropy or token probabilities, to detect and correct unreliable generations. For example, \citet{Wang2025BeyondReasoning} find that only a small subset of generated tokens exhibit high entropy and that these tokens correlate with critical reasoning transitions, motivating entropy-aware interventions that selectively regulate uncertain decoding steps.

Motivated by their efficiency, recent studies have extended similar ideas to MLLM hallucinations, developing training-free decoding interventions that adjust generation~\citep{Leng2024MitigatingDecoding, Xu2025MitigatingDecoding, Xu2026ThinkingDecoding, Huo2025Self-introspectiveModels}. Representative approaches include contrastive decoding and reasoning-aware decoding, which aim to reduce hallucinations by suppressing hallucination-prone tokens or re-weighting generation based on uncertainty estimates~\citep{Leng2024MitigatingDecoding,Xu2025MitigatingDecoding, Xu2026ThinkingDecoding}. However, these methods largely inherit assumptions from unimodal settings and thus treat hallucination as a token-level or uncertainty-driven phenomenon.

\section{Methodology}
\begin{figure*}[t]
    \centering
    \includegraphics[width=\textwidth]{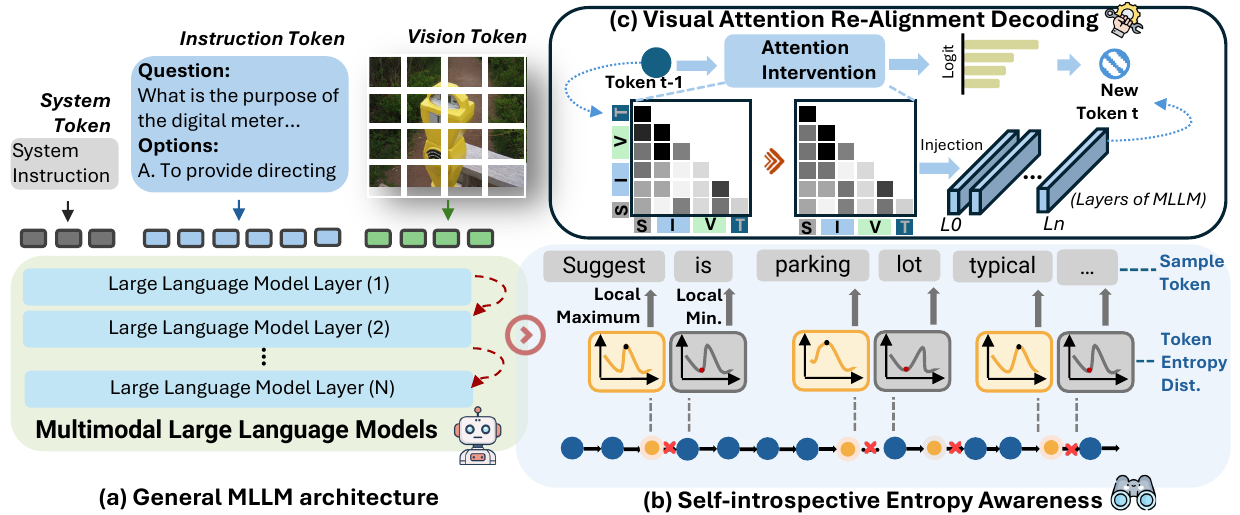}
    \caption{Framework overview of the proposed \textbf{SEED}. The figure presents three parts: (a) A standard MLLM architecture processes system instructions and visual tokens through stacked decoder layers to perform multimodal reasoning.
(b) A self-introspective mechanism monitors token-level entropy dynamics throughout the decoding process, capturing locally important spikes that signal shifts in reasoning dynamics.
(c) When both detection conditions are met, our method performs an attention intervention during decoding to rectify the multimodal interaction. We refer to this process as Visual Attention Re-alignment Decoding.}
    \label{fig:pipeline}
\end{figure*}
This section presents SEED (Self-introspective Entropy-aware Visual Re-alignment Decoding), illustrated in Figure~\ref{fig:pipeline}. SEED is a training-free framework for improving MLLM reasoning that leverages entropy as a lens for monitoring the reasoning trajectory. We introduce a self-introspective mechanism to dynamically re-align visual attention throughout the reasoning trajectory.

\subsection{Problem Formulation}
Given a multimodal large language model (MLLM) $\mathcal{M}_\theta$ parameterized by $\theta$, we consider it as a conditional sequence generator that produces an output token sequence based on both textual and visual inputs. For an image-grounded reasoning task, the input consists of a textual query $x$ and an image $v$. These inputs are jointly encoded into a unified token sequence, which conditions the autoregressive decoding process. The model then generates an output sequence $y=(y_1, y_2,\dots,y_T)$ token by token, following the distribution
\begin{align}
    y_t \sim p_\theta(\cdot \mid x, v, y_{<t}) \propto \mathrm{softmax}(f_\theta(\cdot \mid x, v, y_{<t})),
\end{align}
where $y_{<t}$ denotes the previously generated tokens, and $f_\theta$ represents the unnormalized logits over the vocabulary. Under this formulation, multimodal reasoning can be viewed as a sequential decision process over tokens. Hallucination occurs when the generated sequence $y$ deviates from the visual context in $v$. Rather than treating hallucination as an isolated prediction error, we study it as a dynamic phenomenon emerging along the generation trajectory. To this end, the token-level entropy is recorded to characterize the model's decoding behavior at each step:
\begin{align}
H_t = - \sum_{w \in \mathcal{V}} p_\theta(w \mid x, v, y_{<t}) \log p_\theta(w \mid x, v, y_{<t}).
\label{eq:entropy}
\end{align}

\subsection{Self-introspective Entropy Dynamics Awareness}
Given the entropy sequence $\{H_t\}$ obtained from Eq.~\eqref{eq:entropy}, we maintain an online estimate of the entropy trajectory during decoding. Motivated by the association between hallucination onset and characteristic entropy transitions, we identify two criteria that jointly characterize unstable reasoning states. A multimodal reasoning trajectory is intervened upon when it satisfies both:

\textit{(1) Relative Low-entropy Regime.} At a decoding step within the monitoring window, we identify relative overconfidence when the current entropy falls below the mean of the preceding trajectory:
\begin{equation}
    {H}_t < \tau_{\text{low}}, \qquad t \in [t_0, t_1],
\end{equation}
where $t_0 < t_1$ exclude the initial decoding steps and reserve sufficient steps for completing the CoT process. The threshold is adapted to the current trajectory as $\tau_{\text{low}}=\frac{1}{t-1}\sum_{i<t}H_i$.

\textit{(2) Local Entropy Spike.} When a sharp transient increase in entropy exceeds a high threshold, the MLLM is considered to be experiencing instability arising from cross-modal conflicts. We detect this by:
\begin{equation}
\Delta H_t = H_t - H_{t-1} > \tau_{\text{rise}},
\end{equation}
where $\tau_{\text{rise}}$ is a hyperparameter. Jointly, we characterize hallucination-prone states as those in which the model operates under an overall overconfident convergence regime while simultaneously exhibiting localized high uncertainty. Such a global-local mismatch indicates a breakdown in multimodal coordination and substantially increases the risk of erroneous generation.

\subsection{Visual Attention Re-Alignment Decoding}
Once a hallucination-prone state is detected, instead of performing token-level interventions at the final generation layer, we carry out an attention-level intervention designed to restore visual grounding and better coordinate multimodal interactions without modifying the MLLM's architecture. The implementation consists of two phases.

\paragraph{Prefill phase.} Given the input vision tokens $v = \{v_1, v_2,\dots, v_n\}$, to decouple visual saliency from textual reasoning, we hook the self-attention over the $n$ image tokens from the final layer of the vision encoder and set it as the reference visual importance map:
\begin{align}
\mathcal{A}_{\text{ref}} = \text{MEAN}_{\text{heads}}\left(\sum_i^h \mathcal{A}_i\right)\in \mathbb{R}^{n \times n}, \quad \text{where}~\{\mathcal{A}_i\}\in \mathbb{R}^{h\times n \times n }.
\end{align}
We derive a visual guidance representation $e_{\text{vis}}$ by using the reference attention weights to aggregate the visual embeddings $V$:
\begin{align}
    e_{\text{vis}}=\sum_{j=1}^n\left(\frac{\mathcal{A}_{\text{ref}, j}}{\sum\mathcal{A}_{\text{ref}}}\right)\cdot V_j.
\end{align}
Therefore, $e_{\text{vis}}$ compresses the high-dimensional visual information into a single vector that prioritizes salient tokens over ``background'' tokens, while effectively maintaining the original semantic structure of the input.

\paragraph{Injection phase.} We inject the re-aligned representation at the first layer of the language model to influence the hidden states at the earliest possible stage. The hidden state $\mathbf{h}_t$ is modified via a residual addition: $\mathbf{h}'_t =\mathbf{h}_t +\alpha\cdot e_{\text{vis}}$, where $\alpha$ is the intervention strength hyperparameter that controls the degree of visual re-alignment. For Qwen3-VL, $V$ denotes the layer-0 hidden states at the visual-token positions, which already lie in the residual-stream space. For Llama-3.2-Vision, whose visual features enter through cross-attention, $e_{\text{vis}}$ is taken from the first cross-attention layer and rescaled to the root-mean-square magnitude of $\mathbf{h}_t$ before injection, so that $\alpha$ has a consistent meaning across architectures.

\section{Experiments}
\subsection{Experimental Setup}\label{setup}
\paragraph{Benchmarks.} Our method is evaluated across 6 VLM benchmarks covering 9 evaluation settings, grouped into two categories: vision-language reasoning and visual hallucination tasks. 1)~\textit{Vision-language reasoning tasks} (VStar, MMVP, VisOnly, and M$^3$CoT) mainly assess a model's ability to perform semantically grounded reasoning by jointly integrating visual evidence, world knowledge and linguistic understanding, including compositional reasoning, contextual interpretation, and cross-modal inference.
2)~\textit{Visual hallucination tasks} include POPE~\citep{Li2023EvaluatingModels} and HallusionBench~\citep{Guan2024HallusionBench:Models}, two specialized benchmarks designed to assess whether a model's responses remain faithful to the provided context (e.g., images), thereby directly evaluating models' reliability. More details about the benchmarks can be found in Appendix~\ref{baselines}.

\paragraph{Baseline models.} We adopt three popular MLLMs (Qwen3-VL-8B-Instruct, Qwen3-VL-4B-Instruct~\citep{Bai2025Qwen3-VLReport}, and Llama-3.2-11B-Vision~\citep{Grattafiori2024TheModels}) as backbones to verify that our proposed method generalizes effectively across models of different scales and architectures. For performance and efficiency comparison, we also evaluate against existing training-free hallucination mitigation and reasoning enhancement methods, including VCD~\citep{Leng2024MitigatingDecoding}, MemVR~\citep{Zou2025LookModels} and OPERA~\citep{Huang2024OPERA:Retrospection-Allocation}. We keep generation hyperparameters, including max\_tokens and temperature, consistent for fair comparison. More details about how these methods are designed to mitigate hallucinations can be found in Appendix~\ref{baselines}.

\paragraph{Implementation details and hyperparameter protocol.}\label{implementation} For all experiments, we monitor the entropy trajectory from decoding step 5 ($t_0$) to step 60 ($t_1$). We set the entropy-rise threshold $\tau_{\text{rise}}{=}0.25$ and inject visual features with strength $\alpha{=}0.5$ for two decoding steps after each trigger. The monitoring window is motivated by an analysis of 270 VQA trajectories, which shows that hallucinated spans concentrate in the first 20\% of generation (Appendix~\ref{app:early}). Following prior training-free methods~\citep{Zou2025LookModels}, we select $\tau_{\text{rise}}$ and $\alpha$ on a small held-out subset of M$^3$CoT. We then fix all intervention hyperparameters across backbones and benchmarks. This tune-once configuration avoids per-setting optimization and simplifies the application of SEED. Section~\ref{sec:ablation} reports the corresponding sensitivity analysis. The maximum generation length is 512 tokens, and all inference is conducted on four NVIDIA RTX 5090 GPUs with 32GB VRAM.

\subsection{Main Experimental Results}
Tables~\ref{exp1} and~\ref{exp2} present a comprehensive comparison of zero-shot performance across multiple multimodal reasoning benchmarks and specialized hallucination benchmarks under varying model scales and decoding strategies. The core insights from the main experiments are summarized below.

\textbf{Consistent accuracy uplift on comprehensive benchmarks.} Compared to other baseline models, SEED delivers consistently larger and more stable performance gains across multiple benchmarks and baseline models, achieving average improvements of 3.2\%--3.7\% depending on the backbone and improving over the base model in 13 of 15 settings (vs.\ 9/15 for VCD and 3/14 for MemVR). In particular, MemVR and VCD often gain improvements on MMVP and VisOnly but fail consistently on M$^3$CoT, which requires longer and more complex reasoning. In contrast, SEED maintains robust improvements across different models and evaluation cases.

\textbf{Reduced hallucination.} SEED demonstrates a specialized capability in mitigating hallucinations in MLLMs. On two hallucination-focused benchmarks, POPE and HallusionBench, SEED improves the Qwen3-VL-8B and Qwen3-VL-4B backbones by 2.5\% and 3.6\% on average, respectively. These results support our hypothesis that temporal entropy dynamics can serve as effective indicators of emerging hallucination-prone states, enabling timely intervention during reasoning.

\textbf{Failure analysis.} SEED remains net positive on Llama-3.2-11B overall, improving average performance by 3.15 points on reasoning benchmarks and 0.28 points on hallucination benchmarks. However, it decreases accuracy on MMVP and POPE by 0.2--1.8 points. On Llama POPE, VCD degrades substantially across all three splits (by 5.1--8.6 points), whereas MemVR and SEED exhibit smaller and broadly comparable declines; their absolute scores differ by at most 0.8 points. This cross-method pattern suggests that the localized weakness is not specific to SEED. Two factors may contribute. First, Llama-3.2-Vision uses an earlier cross-attention-based design that injects visual information through a separate pathway, so $e_{\text{vis}}$ requires scale alignment before being added to the textual residual stream. Second, POPE's short binary answers amplify small token-level perturbations. A paired transition analysis on Llama POPE-Adversarial (Appendix~\ref{app:failure}) supports the latter explanation: the number of false-positive answers is nearly unchanged (462$\rightarrow$467), whereas invalid-format outputs increase (610$\rightarrow$646). Thus, the net drop stems mainly from failures to produce responses in the required yes/no format rather than increased object hallucination.

\begin{table*}[t]
\centering
\small
\caption{Comparisons across general reasoning benchmarks. The summary column reports the macro-average change in percentage points and the number of strictly positive changes over available settings. Base MLLMs use the default CoT setting.}
\setlength{\tabcolsep}{4pt}
\begin{tabular}{l|lllll|c}
\toprule
\multirow{2}{*}{\textbf{Method}}
& \multicolumn{5}{l|}{\textbf{Multimodal Reasoning \& Understanding}}
& \multirow{2}{*}{\makecell{\textbf{Summary}\\$\Delta$ (\%) / improved}} \\
\cmidrule(lr){2-6}
& \textbf{VStar$\uparrow$}  & \textbf{MMVP$\uparrow$} & \textbf{VisOnly$_s$$\uparrow$} &
\textbf{VisOnly$_r$$\uparrow$}&\textbf{M$^3$CoT$\uparrow$} \\
\midrule

Qwen3-VL-8B
& 73.82  & 76.23 & 42.20 &54.60&78.21 & -- \\
+ VCD
& 54.45(\textcolor{Gray}{-19.4})  & 80.00(\textcolor{Green}{+3.8}) & 42.57(\textcolor{Green}{+0.4}) & 55.71(\textcolor{Green}{+1.1})&77.68(\textcolor{Gray}{-0.5}) & $-2.93$ / 3/5 \\
+ MemVR
&74.87(\textcolor{Green}{+1.0})  & 80.67(\textcolor{Green}{+4.4}) & 41.86(\textcolor{Gray}{-0.3}) &54.44(\textcolor{Gray}{-0.2})&77.68(\textcolor{Gray}{-0.5}) & $+0.89$ / 2/5 \\
\cellcolor{mycolor}+ \textbf{SEED (Ours)}
& \cellcolor{mycolor}83.25(\textcolor{Green}{+9.4})  & \cellcolor{mycolor}76.33(\textcolor{Green}{+0.1}) & \cellcolor{mycolor}47.00(\textcolor{Green}{+4.8}) &\cellcolor{mycolor}54.37(\textcolor{Gray}{-0.2})& \cellcolor{mycolor}79.97(\textcolor{Green}{+1.8}) & \cellcolor{mycolor}\textbf{+3.17 / 4/5} \\
\midrule

Qwen3-VL-4B
& 68.06  & 76.33 & 42.14 &52.22& 76.79 & --\\
+ VCD
& 54.97(\textcolor{Gray}{-13.1})  & 78.67(\textcolor{Green}{+2.3}) & 43.25(\textcolor{Green}{+1.1})& 52.73(\textcolor{Green}{+0.5})&75.28(\textcolor{Gray}{-1.5}) & $-2.13$ / 3/5\\
+ MemVR
&66.49(\textcolor{Gray}{-1.6})  & 76.33(\textcolor{Gray}{-0.0}) & 42.14(\textcolor{Gray}{-0.0}) &54.22(\textcolor{Green}{+2.0})&76.08(\textcolor{Gray}{-0.7}) & $-0.06$ / 1/5 \\
\cellcolor{mycolor}+ \textbf{SEED (Ours)}
& \cellcolor{mycolor}76.76(\textcolor{Green}{+8.7}) & \cellcolor{mycolor}77.33(\textcolor{Green}{+1.0}) & \cellcolor{mycolor}45.57(\textcolor{Green}{+3.4}) &\cellcolor{mycolor}56.42(\textcolor{Green}{+4.2})& \cellcolor{mycolor}78.03(\textcolor{Green}{+1.2}) & \cellcolor{mycolor}\textbf{+3.71 / 5/5} \\
\midrule

Llama-3.2-11B
& 50.26 & 62.33 & 25.00 & 33.40 & 64.66 & -- \\
+ VCD
& 42.41(\textcolor{Gray}{-7.8}) & 63.67(\textcolor{Green}{+1.3}) &27.14(\textcolor{Green}{+2.1}) & 33.91(\textcolor{Green}{+0.5}) &63.13(\textcolor{Gray}{-1.5}) & $-1.08$ / 3/5\\
+ MemVR
&40.31(\textcolor{Gray}{-9.9})  & 60.30(\textcolor{Gray}{-2.0}) & 21.57(\textcolor{Gray}{-3.4})  & 27.78(\textcolor{Gray}{-5.6}) & - & $-5.26$ / 0/4\\
\cellcolor{mycolor}+ \textbf{SEED (Ours)}
& \cellcolor{mycolor}57.61(\textcolor{Green}{+7.3}) & \cellcolor{mycolor}61.33(\textcolor{Gray}{-1.0}) & \cellcolor{mycolor}28.00(\textcolor{Green}{+3.0}) &\cellcolor{mycolor}36.90(\textcolor{Green}{+3.5})& \cellcolor{mycolor}67.58(\textcolor{Green}{+2.9}) & \cellcolor{mycolor}\textbf{+3.15 / 4/5} \\
\bottomrule
\end{tabular}
\label{exp1}
\end{table*}

\begin{table*}[t]
\centering
\small
\caption{Comparisons across hallucination benchmarks. The summary column reports the macro-average change in percentage points and the number of positive changes over available settings. Base MLLMs use the default CoT setting.}
\setlength{\tabcolsep}{4pt}
\begin{tabular}{l|llll|c}
\toprule
\multirow{2}{*}{\textbf{Method}}
& \multicolumn{4}{l|}{\textbf{Multimodal Hallucination}}
& \multirow{2}{*}{\makecell{\textbf{Summary}\\$\Delta$ (\%) / improved}} \\
\cmidrule(lr){2-5}
 & \textbf{HalluBench$\uparrow$} & \textbf{POPE-R$\uparrow$} & \textbf{POPE-P$\uparrow$} & \textbf{POPE-A$\uparrow$} \\
\midrule

Qwen3-VL-8B
 &58.50 & 90.25&87.53   & 86.42 & -- \\
+ VCD
 & 56.24(\textcolor{Gray}{-2.3}) & 91.67(\textcolor{Green}{+1.4}) & 87.03(\textcolor{Gray}{-0.5})& 86.30(\textcolor{Gray}{-0.1}) & $-0.37$ / 1/4 \\
+ MemVR
& 55.87(\textcolor{Gray}{-2.6}) & 90.63(\textcolor{Green}{+0.4}) & 88.12(\textcolor{Green}{+0.6}) &86.85(\textcolor{Green}{+0.4}) & $-0.31$ / 3/4 \\
\cellcolor{mycolor}+ \textbf{SEED (Ours)}
 & \cellcolor{mycolor}59.28(\textcolor{Green}{+0.8}) & \cellcolor{mycolor}97.39(\textcolor{Green}{+7.1}) & \cellcolor{mycolor}90.94(\textcolor{Green}{+3.4}) & \cellcolor{mycolor}85.07(\textcolor{Gray}{-1.3}) & \textbf{\cellcolor{mycolor}+2.49 / 3/4} \\
\midrule

Qwen3-VL-4B
 & 54.80 & 90.92 & 88.64 & 87.96 & -- \\
+ VCD
 & 53.84(\textcolor{Gray}{-1.0}) & 90.60(\textcolor{Gray}{-0.3}) & 87.87(\textcolor{Gray}{-0.8}) &87.37(\textcolor{Gray}{-0.6}) & $-0.66$ / 0/4 \\
+ MemVR
 & 55.27(\textcolor{Green}{+0.5}) & 90.77(\textcolor{Gray}{-0.2}) & 87.69(\textcolor{Gray}{-1.0}) &88.39(\textcolor{Green}{+0.4}) & $-0.05$ / 2/4\\
\cellcolor{mycolor}+ \textbf{SEED (Ours)}
 & \cellcolor{mycolor}54.84(\textcolor{Green}{+0.0}) & \cellcolor{mycolor}98.45(\textcolor{Green}{+7.5})& \cellcolor{mycolor}92.84(\textcolor{Green}{+4.2}) & \cellcolor{mycolor}90.43(\textcolor{Green}{+2.5}) & \cellcolor{mycolor}\textbf{+3.56 / 4/4} \\
\midrule

Llama-3.2-11B
& 42.38 & 91.33 & 88.02 &85.04 & -- \\
+ VCD
 & 43.66(\textcolor{Green}{+1.3}) & 86.23(\textcolor{Gray}{-5.1}) & 80.80(\textcolor{Gray}{-7.2}) &76.43(\textcolor{Gray}{-8.6}) & $-4.91$ / 1/4 \\
+ MemVR
& 43.16(\textcolor{Green}{+0.8}) & 91.02(\textcolor{Gray}{-0.3}) & 87.86(\textcolor{Gray}{-0.2}) & 83.83(\textcolor{Gray}{-1.2}) & $-0.23$ / 1/4\\
\cellcolor{mycolor}+ \textbf{SEED (Ours)}
 & \cellcolor{mycolor}46.42(\textcolor{Green}{+4.0}) & \cellcolor{mycolor}91.10(\textcolor{Gray}{-0.2}) & \cellcolor{mycolor}87.10(\textcolor{Gray}{-0.9}) & \cellcolor{mycolor}83.27(\textcolor{Gray}{-1.8}) & \cellcolor{mycolor}\textbf{+0.28 / 1/4} \\
\bottomrule
\end{tabular}
\label{exp2}
\end{table*}

\subsection{Ablation Studies}\label{sec:ablation}
\paragraph{Variants of attention intervention strategies.}
As shown in Table~\ref{tab:ablation1}, we further analyze multiple variants of visual intervention strategies. The results indicate that performance is highly sensitive to the design of attention intervention, highlighting the importance of properly coordinating multimodal attention during reasoning. Specifically, ``\texttt{Equal attention}'' resets the text-to-image attention map to equal weights; ``\texttt{Truncated SVD Project}'' performs SVD on the attention key matrix and retains only the top-$k$ principal components to form a low-rank orthogonal subspace, so that the MLLM receives compressed, important visual information for re-alignment; ``\texttt{ViT attention intervene}'' randomly perturbs the attention distribution in the ViT layers to isolate the effect of visual information. More details of the variants can be found in Appendix~\ref{variants}.

Although some variants (e.g., ``\texttt{Equal attention}'') effectively improve models' accuracy by 2.0\%, they often sacrifice inference efficiency with 4$\times$ longer inference time, while efficient variants suffer severe performance drops. In contrast, SEED achieves a favorable balance between effectiveness and computational efficiency across various backbones. End-to-end, SEED increases latency by less than 5\% (3.30 to 3.45 ms/token), compared with 54\% for VCD, 157\% for MemVR, and 333\% for OPERA; full results appear in Appendix~\ref{tab:efficiency}.

\begin{table}[t]
\centering
\small
\caption{Comparison of attention intervention strategies across backbones on M$^3$CoT~\citep{Chen2024M3CoT:Chain-of-Thought}. Each backbone is evaluated in terms of accuracy and efficiency (instance/s). All hyperparameters are kept consistent. The best and second-best results are highlighted in green and blue, respectively.}
\label{tab:ablation1}
\setlength{\tabcolsep}{6pt}
\renewcommand{\arraystretch}{1.2}
\begin{tabular}{lll|ll}
\toprule
\textbf{Attention Intervention Strategy}
& \multicolumn{2}{l|}{\textbf{Qwen3-VL-8B}}
& \multicolumn{2}{l}{\textbf{Qwen3-VL-4B}} \\
\cmidrule(lr){2-3} \cmidrule(lr){4-5}
& \textbf{Accuracy} $\uparrow$
& \textbf{Efficiency} $\uparrow$
& \textbf{Accuracy} $\uparrow$
& \textbf{Efficiency} $\uparrow$ \\
\midrule
None (baseline)
& 78.21 & 0.20
& 76.79& 0.22 \\
+ Equal attention
& \cellcolor{green!15}80.24(+2.0) & 0.05($\times$0.25)
& \cellcolor{blue!15}78.02(+1.2) & 0.07($\times$0.32)\\
+ Truncated SVD Project
& 54.28(-23.9) & \cellcolor{blue!15}0.17($\times$0.85)
& 50.23(-26.6) & \cellcolor{blue!15}0.18($\times$0.82) \\
+ ViT attention intervene
& 60.32(-17.9) & 0.15($\times$0.75)
& 55.36(-21.4) & 0.16($\times$0.73) \\
\midrule
+ SEED (Ours)
& \cellcolor{blue!15}\textbf{79.97}(+1.8) &\cellcolor{green!15}0.19($\times$0.95)
& \cellcolor{green!15}\textbf{78.03}(+1.2) & \cellcolor{green!15}0.21($\times$0.95) \\
\bottomrule
\end{tabular}
\end{table}

\paragraph{Hyperparameter sensitivity.}
Since SEED uses a single hyperparameter configuration across all models and benchmarks, hyperparameter sensitivity is examined in this section. On M$^3$CoT with Qwen3-VL-8B, no tested setting across $\alpha \in [0.4, 0.7]$, $\tau_{\text{rise}} \in [0.225, 0.300]$, and $t_1 \in [40, 70]$ falls below the base model, and all but $\alpha{=}0.7$ improve on it (Appendix~\ref{hyper}). Together, the fixed cross-setting protocol and broad stable range indicate that SEED does not depend on narrowly tuned hyperparameter values.

\section{Conclusion}
In this work, we rethink hallucination in MLLMs as a dynamical phenomenon emerging from the decoding trajectory. By interpreting entropy as a structured signal of multimodal interaction, we show that hallucinations are closely tied to two coupled behaviors: a globally overconfident regime that reflects premature reliance on ineffective reasoning, and locally unstable entropy fluctuations that expose conflicts in cross-modal reasoning. Motivated by these findings, we propose an entropy-aware self-introspective framework that uses entropy as a real-time diagnostic and intervention signal for targeted visual re-alignment during critical reasoning steps. This shifts hallucination mitigation from passive uncertainty modeling to active control over multimodal interaction dynamics. Empirically, SEED improves 21 of 27 model--benchmark settings, validating the proposed approach, as well as the broader premise that structured decoding signals can help identify and partially correct unreliable reasoning states. Looking forward, promising directions include query-dependent visual representations, and adaptive intervention strengths, which may further improve SEED's robustness across model families. We believe that such entropy-centric analysis provides a foundation for more controllable and interpretable multimodal reasoning systems.

\subsection*{AI Use Statement}
Generative AI tools were used to refine the writing of this manuscript and to assist with code development and debugging. All research design, methodological decisions, experimental procedures, and interpretation of results were executed by the authors. All AI-assisted text and code were reviewed, verified, and, where applicable, tested by the authors. The authors take full responsibility for the final content of this work.

\bibliography{references}
\bibliographystyle{iclr2027_conference}

\newpage
\appendix
\section{More Preliminary Experiment Results}\label{preliminary}
\subsection{Localized entropy spikes mark hallucination onset}
\begin{figure*}
    \centering
    \includegraphics[width=\textwidth]{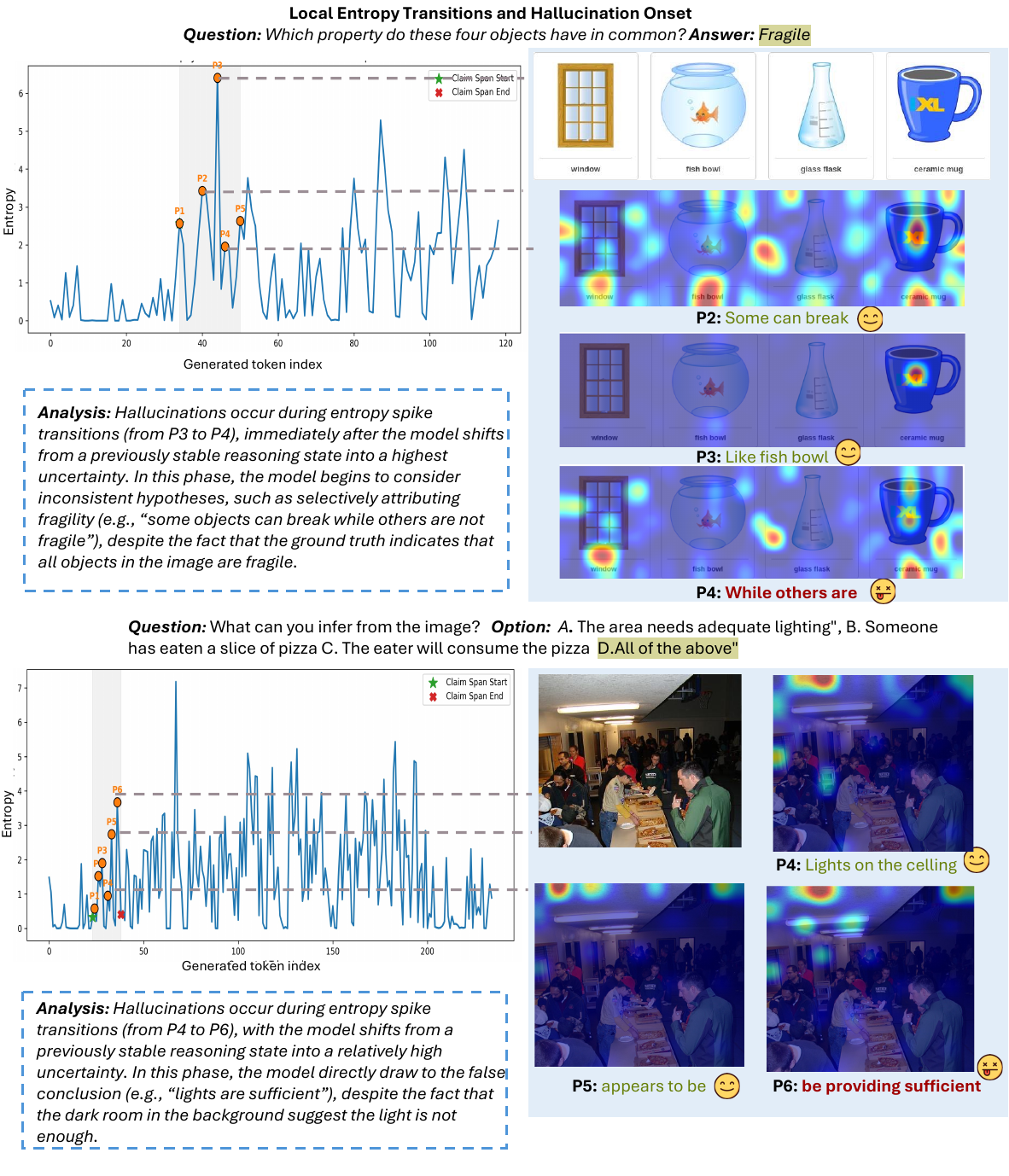}
    \caption{More examples illustrating hallucination emergence during MLLM reasoning. The left panel shows entropy over generated tokens, while the right panel visualizes the corresponding visual attention patterns at selected reasoning steps. The gray region denotes the span of incorrect generation. Hallucinated reasoning coincides with regions containing sharp entropy spikes and increasingly strong yet diffuse visual attention.}
    \label{fig:preliminary2_more}
\end{figure*}

As illustrated in Figure~\ref{fig:preliminary2_more}, we provide more cases to understand how hallucinations emerge. These cases show that hallucination is closely coupled with sharp transitions in the model’s local entropy. As shown in the figure, hallucinations typically manifest during entropy spike transitions (e.g., the $P3 \to P4$ transition in the top case and $P4 \to P6$ in the bottom case), where the model shifts from a stable reasoning state into a regime of high uncertainty. During these high-entropy phases, the MLLM begins to generate inconsistent content. For the top example, the MLLM selectively attributes fragility to only a subset of objects, despite clear visual evidence to the contrary. 

Furthermore, we observe a distinct temporal coupling between these entropy spikes and the diffuseness of visual attention. While successful reasoning is characterized by a rapid convergence and stabilization of visual attention, hallucinatory trajectories exhibit a prolonged yet ineffective visual dependency. The spikes therefore provide a useful marker of a breakdown in modality integration: the model fails to effectively map visual tokens to its linguistic output, leading to the generation of ``confident'' but factually ungrounded falsehoods. This alignment suggests that entropy does not merely reflect token-level static output uncertainty, but also tracks changes in multimodal information interaction.
\subsection{Hallucinations mainly occur at the early stage}\label{app:early}
\begin{figure*}
    \centering
    \includegraphics[width=\textwidth]{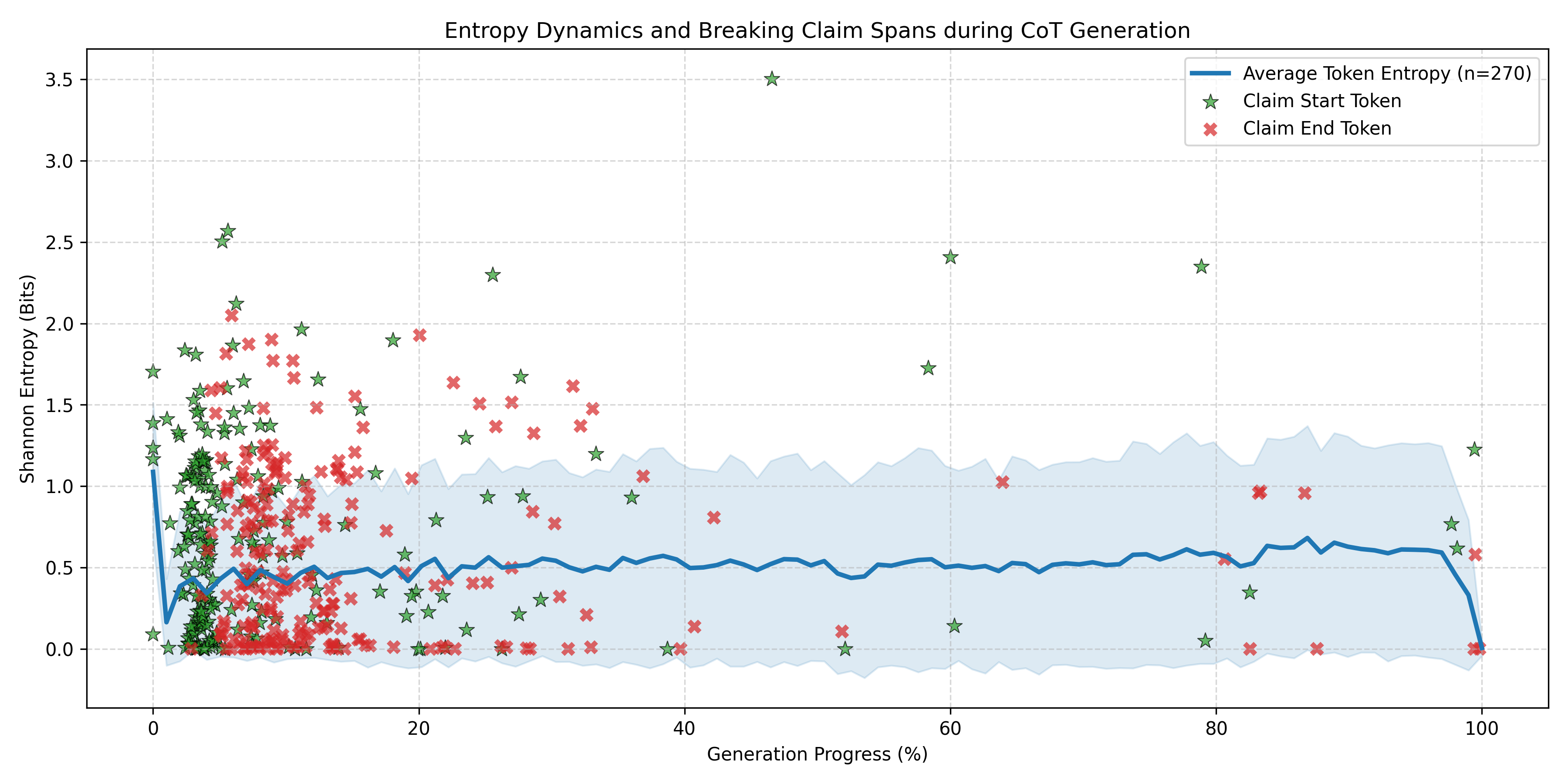}
    \caption{Statistical analysis of wrong reasoning span sizes. The reasoning processes of 270 VQA samples are recorded. Results show that wrong reasoning claims concentrate in the first 20\% of the generation process.}
    \label{fig:claim span}
\end{figure*}
To characterize when erroneous reasoning emerges, we analyze 270 VQA reasoning trajectories and record the start and end tokens of each erroneous span (Figure~\ref{fig:claim span}). These spans are concentrated within the first 20\% of generation and typically extend for approximately 50--70 tokens within the maximum length of 512 tokens. Their early occurrence, together with their association with local entropy spikes in Figure~\ref{fig:preliminary2_more}, motivates SEED's focus on early decoding. When an entropy trigger is detected, SEED injects the attention-weighted visual feature $e_{\text{vis}}$ at Layer~0 to reinforce visual grounding from the start of the language-model computation.
\subsection{Population-level entropy analysis}\label{app:population}
To complement the case studies above, we analyze 1,228 randomly sampled VQA trajectories generated by Qwen3-VL-8B, of which 763 contain hallucinations and 465 are correct and serve as controls (following the original accuracy distribution). For each hallucinated trajectory, we identify the first token at which hallucinated content appears and align it to relative position $p{=}0$. We then compute the mean token-to-token entropy change $\Delta H = H_t - H_{t-1}$ in a $\pm 20$-token window around the onset. Since correct trajectories have no hallucination onset, $p{=}0$ is sampled uniformly at random for them. Table~\ref{tab:onset} reports the resulting profile. At the onset, the hallucination group exhibits a mean entropy jump of 0.245, more than $20\times$ larger than that of the control group (0.012), followed by a sharp drop at $p{=}{+}1$. A two-sided Mann--Whitney U test on the per-response $\Delta H$ at $p{=}0$ gives $p = 4\times 10^{-4}$.

We further compare three trajectory-level statistics (Table~\ref{tab:population} in the main text): the length of the longest low-entropy run (low-entropy duration), the fraction of steps whose $\Delta H$ exceeds $\tau_{\text{rise}}$ (spike frequency), and the maximum $\Delta H$ within a trajectory. Hallucinated trajectories show significantly longer low-entropy durations ($p<10^{-35}$), more frequent spikes ($p<10^{-9}$), and larger maximum entropy changes ($p<10^{-16}$), consistent with the two patterns identified in Section~2. These results establish a robust statistical association; they do not by themselves establish that entropy spikes \emph{cause} hallucinations, and SEED only relies on them as a detection signal.

\begin{table}[h]
\centering
\small
\caption{Mean token-to-token entropy change $\Delta H$ aligned to the hallucination onset ($p{=}0$). For ranges, the mean over the range (or its min--max) is reported.}
\label{tab:onset}
\setlength{\tabcolsep}{3.5pt}
\resizebox{\textwidth}{!}{%
\begin{tabular}{lcccccccccc}
\toprule
$p$ & $-20{\sim}-11$ & $-5$ & $-4$ & $-3$ & $-2$ & $-1$ & $\mathbf{0}$ & $+1$ & $+2{\sim}+5$ & $+6{\sim}+20$ \\
\midrule
Hallucinated & 0.0020 & 0.0560 & 0.0011 & 0.0188 & $-$0.0423 & 0.0065 & \textbf{0.2447} & $-$0.1728 & $-$0.0227$\sim$0.0081 & $-$0.0025 \\
Correct & 0.0007 & 0.0119 & 0.0140 & $-$0.0010 & $-$0.0109 & 0.0047 & 0.0117 & $-$0.0125 & $-$0.0404$\sim$0.0265 & 0.0007 \\
\bottomrule
\end{tabular}}
\end{table}

\section{Experiment Details}
\subsection{Benchmarks}\label{baselines}
Our method is evaluated across multiple VLM benchmarks, grouped into two main categories: general vision-language understanding and visual hallucination task.  1) \textit{General multimodal reasoning task} assesses how MLLM can effectively incorporate information and use it as the reference for complex reasoning tasks. There are four benchmarks in this category: 
\begin{itemize}
    \item $\text{M}^3\text{CoT}$~\citep{Chen2024M3CoT:Chain-of-Thought} specifically addresses three critical limitations found in existing MCoT evaluations: the absence of visual modal reasoning, the prevalence of single-step visual reasoning, and a lack of domain diversity. By facilitating a multi-domain, multi-step scenario, $\text{M}^3\text{CoT}$ provides a more rigorous environment for testing how MLLMs leverage knowledge from both textual and visual modalities for \textit{complex (long-horizon)} and step-by-step reasoning. In our experiments, we use its validation set for testing, which contains 1,110 VQA instances.
    \item MMVP~\citep{Tong2024EyesLLMs} is a compact benchmark that contains 300 image-question pairs. It is designed to evaluate vision-language models on ``CLIP-blind'' visual features. Higher performance on MMVP indicates the MLLMs can genuinely utilize visual information rather than relying on language priors.
    \item VisOnly (Synthetic and Real splits)~\citep{Kamoi2025VisOnlyQA:Information} evaluates MLLMs on 12 geometric perception tasks, and reveals that base MLLMs often cannot accurately perceive basic geometric information in images. The dataset contains a synthetic split (700 questions) and a real split (900 questions).
    \item VStar~\citep{Wu2023V:LLMs} proposes a visual search mechanism together with a benchmark to address a critical limitation in current MLLMs: the inability to selectively process visual details in high-resolution or crowded images.
    
\end{itemize}

 2) \textit{Visual hallucination task.} In addition to assessing general MLLM capabilities, we conduct evaluations on two specialized benchmarks, where our method demonstrates strong effectiveness in mitigating hallucination:
 \begin{itemize}
     \item POPE~\citep{Li2023EvaluatingModels} is a diagnostic benchmark for probing object hallucination in MLLMs via binary (yes/no) questions about object presence. It comprises 3,000 questions constructed under different sampling strategies (e.g., random, popular, adversarial) to systematically test whether models can correctly assert the existence of certain objects.
     \item HallusionBench~\citep{Guan2024HallusionBench:Models} comprises 346 images paired with 1,129 questions, all meticulously crafted by human experts. It is specifically designed to evaluate multimodal hallucination across diverse visual reasoning scenarios. 
 \end{itemize}
\subsection{Baseline models}
\begin{itemize}
    \item \textbf{Base Models.} We select Qwen3-VL-4B-Instruct, Qwen3-VL-8B-Instruct~\citep{Bai2025Qwen3-VLReport}, and Llama-3.2-11B-Vision~\citep{Grattafiori2024TheModels} as our primary backbones to ensure diversity in both model scale and architectural design. Specifically, Qwen3-VL models represent strong \textbf{decoder-only self-attention} architectures at different parameter scales (4B vs 8B), enabling us to analyze scaling effects on our proposed method. In contrast, Llama-3.2-11B-Vision adopts a \textbf{cross-attention-based design}, providing a complementary paradigm for multimodal fusion. This combination allows us to comprehensively evaluate the generality of our method across distinct model architectures and capacities.
    \item \textbf{Other Baseline Models.} We also consider alternative training-free decoding strategies that aim at alleviating MLLMs' hallucinations, including VCD~\citep{Leng2024MitigatingDecoding} and MemVR~\citep{Zou2025LookModels}. Specifically, VCD mitigates hallucination by contrasting model predictions under visual-conditioned and text-only contexts during decoding, reweighting candidate tokens to favor those consistent with visual evidence. MemVR improves visual reasoning by maintaining a visual memory bank of vision representations, enabling the model to retrieve and reinforce previously grounded information to ensure consistency across multi-step reasoning.
\end{itemize}
\subsection{Reproducibility}\label{reproduce} Besides $\text{M}^3\text{CoT}$~\citep{Chen2024M3CoT:Chain-of-Thought}, we employ OpenCompass VLMEvalKit~\citep{Duan2024VLMEvalKit:Models} to conduct evaluations for fair comparison. We follow the official evaluation document of $\text{M}^3\text{CoT}$ for its own evaluation. During the inference of baseline models (MemVR and VCD), we follow their prompt instructions if they are available. 
\subsection{Computation costs}
As summarized in Table~\ref{tab:efficiency}, SEED introduces negligible inference overhead ($\sim$5\% increase in time cost per 100 tokens) while improving performance over the greedy baseline.

\begin{table}[t]
\centering
\caption{Performance comparison of SOTA methods and SEED in latency, throughput, and time cost. All methods are built on the Qwen3-VL-8B-Instruct backbone. The experiment is conducted on $\text{M}^3\text{CoT}$ and all models are prompted to generate CoT. Note that all results are evaluated under an inference acceleration framework.}
\label{tab:efficiency}
\setlength{\tabcolsep}{4pt}
\renewcommand{\arraystretch}{1.2}

\begin{tabular}{lccc}
\toprule
\textbf{Method} 
& \makecell{\textbf{Latency $\downarrow$}\\(ms/token)} 
& \makecell{\textbf{Throughput $\uparrow$}\\(instance/s)} 
& \makecell{\textbf{Time Cost $\downarrow$}\\(min)} 
\\
\midrule

Greedy 
& 3.30 {\small ($\times$1.00)} 
& 0.941 {\small ($\times$1.00)} 
&20.0 {\small ($\times$1.00)} 
 \\

OPERA 
& 14.28 {\small ($\times$4.33)} 
& 0.224 {\small ($\times$0.24)} 
& 85.0 {\small ($\times$4.25)} 
 \\

VCD 
& 5.08 {\small ($\times$1.54)} 
& 0.598 {\small ($\times$0.63)} 
& 31.5 {\small ($\times$1.58)} 
 \\

MemVR 
& 8.47 {\small ($\times$2.57)} 
& 0.359 {\small ($\times$0.38)} 
& 52.5 {\small ($\times$2.63)} 
 \\
\midrule
SEED 
& \cellcolor{green!15}3.45 {\small ($\times$1.05)} 
& \cellcolor{green!15}0.897 {\small ($\times$0.95)} 
& \cellcolor{green!15}21.0 {\small ($\times$1.05)} 
\\
\bottomrule
\end{tabular}

\end{table}

\subsection{Failure analysis}\label{app:failure}
\paragraph{Paired transition analysis.} To understand when SEED's interventions are harmful, we compare the per-sample predictions of the base model and SEED on the representative failure case, Llama-3.2-11B-Vision on POPE-Adversarial (Table~\ref{tab:transition}). SEED corrects 408 of the 1,137 samples the base model answers incorrectly, while flipping 454 correct answers, a net change of 46 samples. Breaking down this net degradation, the number of false-positive answers (i.e., hallucinated objects) barely changes (462$\rightarrow$467), whereas invalid-format outputs increase from 610 to 646. Hence the drop is mainly associated with instruction-following failures under constrained binary evaluation---specifically, outputs that do not follow the required yes/no format---rather than a systematic increase in object hallucination. Interestingly, persistently wrong samples receive the strongest intervention signals (highest trigger ratio, most trigger events, and longest intervention span) yet remain hard to correct, suggesting that entropy dynamics identify difficult reasoning states even though not all of them are amenable to visual re-alignment. POPE's short trajectories and binary outputs further make it sensitive to local token-level perturbations introduced during intervention.

\begin{table}[h]
\centering
\small
\caption{Paired transition analysis between the base model and SEED on POPE-Adversarial with Llama-3.2-11B-Vision.}
\label{tab:transition}
\setlength{\tabcolsep}{4pt}
\begin{tabular}{llccccc}
\toprule
\textbf{Transition} & \textbf{Interpretation} & \textbf{Count} & \textbf{Rate} & \textbf{Trigger ratio} & \textbf{Avg. triggers} & \textbf{Avg. span} \\
\midrule
Correct $\rightarrow$ Wrong & Harmful intervention & 454 & 15.13\% & 46.26\% & 2.04 & 7.95 \\
Wrong $\rightarrow$ Correct & Successful correction & 408 & 13.60\% & 43.38\% & 2.20 & 8.69 \\
Wrong $\rightarrow$ Wrong & Unresolved error & 729 & 24.30\% & 60.22\% & 2.93 & 11.84 \\
\bottomrule
\end{tabular}
\end{table}

\subsection{Ablation Variants}\label{variants}
Here are the implementation details of attention intervention strategies.
\begin{itemize}
    \item Equal attention: Earlier work~\citep{Fazli2025MitigatingCalibration} shows that even (i.e., uniform) visual attention can help MLLMs alleviate hallucinations. Following this idea, we set up equal attention, which naively re-guides the model to attend to important visual evidence.
    \item Truncated SVD: Following~\citet{Lin2026HulluEdit:Models}, we construct a low-rank visual projection matrix using truncated singular value decomposition (SVD) to constrain attention intervention within a compact visual subspace. Given the visual key matrix $K_v \in \mathbb{R}^{N_v \times d}$, we first perform singular value decomposition:
    \[
    K_v = U \Sigma V^\top .
    \]
    We then retain the top-$k$ principal components from $V$ to obtain a low-rank orthogonal basis $V_k \in \mathbb{R}^{d \times k}$. The visual projection matrix is constructed as
    \[
    P_{\text{vis}} = V_k V_k^\top ,
    \]
    which projects attention intervention into the dominant visual subspace while avoiding explicit matrix inversion. The method is designed to keep the salient visual information and re-inject it into the reasoning process.
    \item ViT attention intervene: Similar to multimodal contrastive methods~\citep{Liu2024MitigatingTuning, Zhou2025MitigatingCausality} that remove images to isolate the impact of language priors from the reasoning, we instead randomly perturb the attention map during the ViT self-attention process so the generated visual embedding will be ineffective. By doing so, we conduct contrastive decoding by actively comparing the results between intervened visual information and normal ones. The experimental results in Table~\ref{tab:ablation1} show that such contrastive decoding still requires careful design of interactions.
    
\end{itemize}
\subsection{Hyperparameter analysis}\label{hyper}
\begin{figure}[h]
\centering
\includegraphics[width=0.62\textwidth]{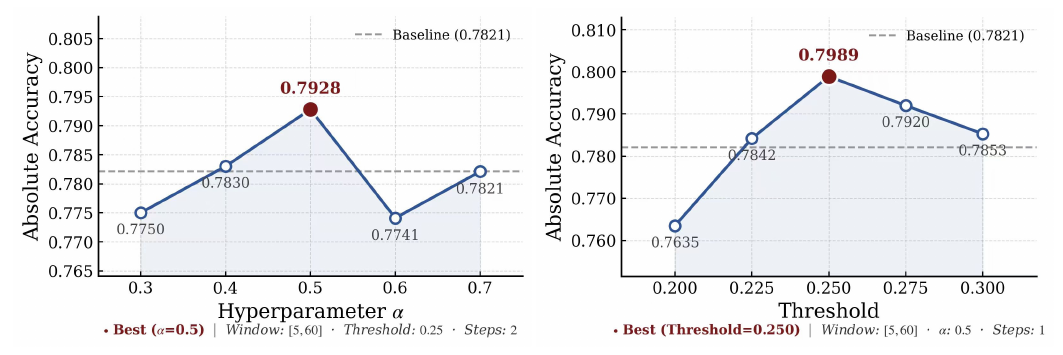}
\caption{Hyperparameter analysis of injection ratios (left) and entropy threshold (right).}
\label{hyper_main}
\end{figure}

\begin{figure*}
    \centering
    \includegraphics[width=0.8\textwidth]{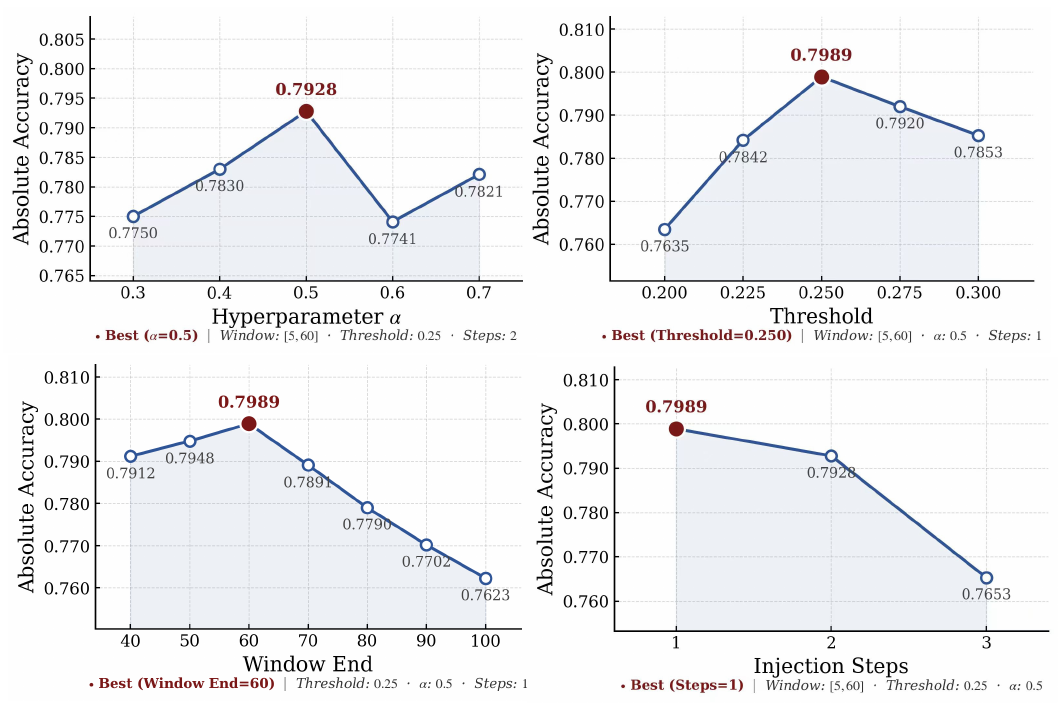}
    \caption{Additional hyperparameter analysis of the entropy threshold, injection ratio, and the start and end steps of the monitoring window.}
    \label{fig:hyper_more}
\end{figure*}

All hyperparameters are selected once and shared across all backbones and benchmarks. Table~\ref{tab:hyper_table} reports the sensitivity of SEED (Qwen3-VL-8B, M$^3$CoT) to the injection coefficient $\alpha$, the trigger threshold $\tau_{\text{rise}}$, and the end of the monitoring window $t_1$, varying one hyperparameter at a time while fixing the others at their default values ($\alpha{=}0.5$, $\tau_{\text{rise}}{=}0.25$, $t_0{=}5$, $t_1{=}60$). No setting falls below the base model, and all but one improve on it. Figure~\ref{hyper_main} visualizes the sensitivity to $\alpha$ and $\tau_{\text{rise}}$, and further results, including the start of the monitoring window, are shown in Figure~\ref{fig:hyper_more}.

\begin{table}[h]
\centering
\small
\caption{Hyperparameter sensitivity on M$^3$CoT with Qwen3-VL-8B (base model: 78.21). Gains over the base model are shown in parentheses.}
\label{tab:hyper_table}
\setlength{\tabcolsep}{5pt}
\begin{tabular}{lcccc}
\toprule
$\alpha$ & 0.40 & 0.45 & \textbf{0.50} & 0.70 \\
 & 78.30 (+0.09) & 78.79 (+0.58) & \textbf{79.89 (+1.68)} & 78.21 (+0.00) \\
\midrule
$\tau_{\text{rise}}$ & 0.225 & \textbf{0.250} & 0.275 & 0.300 \\
 & 78.42 (+0.21) & \textbf{79.89 (+1.68)} & 79.20 (+0.99) & 78.53 (+0.32) \\
\midrule
$t_1$ & 40 & 50 & \textbf{60} & 70 \\
 & 79.12 (+0.91) & 79.48 (+1.27) & \textbf{79.89 (+1.68)} & 78.91 (+0.70) \\
\bottomrule
\end{tabular}
\end{table}
\subsection{Case studies}\label{case study}
\begin{figure*}[h]
    \centering
    \includegraphics[width=\textwidth]{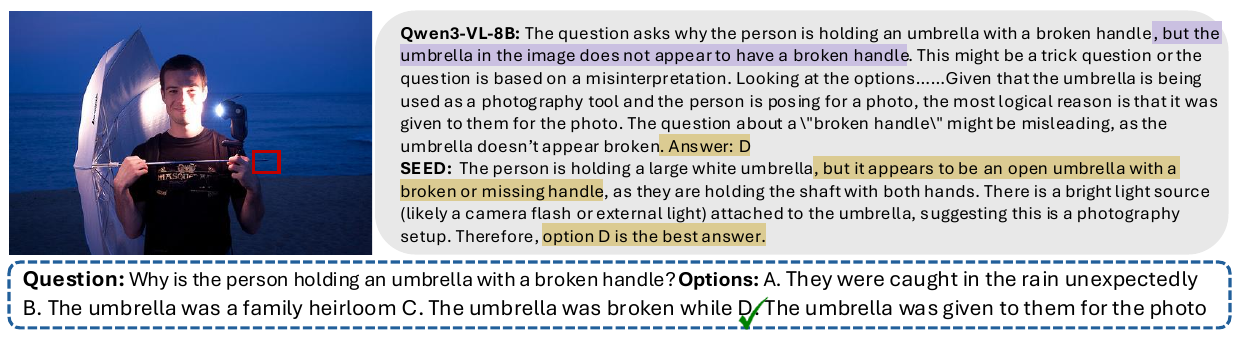}
    \caption{A case study illustrating that SEED not only improves final answer accuracy but also produces more reliable intermediate reasoning steps. Hallucinated content is highlighted in purple, while correct reasoning is highlighted in yellow.}
    \label{fig:case}
\end{figure*}

We further present a case study to investigate the effectiveness of our method and its impact on the entropy trajectory during decoding. As shown in Figure~\ref{fig:case}, although both the baseline Qwen3-VL-8B and SEED yield the same correct final answer, SEED exhibits improved visual grounding and produces more faithful intermediate descriptions. Specifically, Qwen3-VL-8B incorrectly rejects the premise by claiming that \texttt{the umbrella handle is not broken}, which is not supported by the visual evidence, where \texttt{the umbrella handle is obviously missing}. In contrast, SEED better aligns its reasoning with the image and incorporates this detail into a coherent explanation. This suggests that SEED not only preserves answer correctness but also enhances the reliability of the reasoning process, which is more important in long-horizon reasoning applications.

\begin{figure*}
    \centering\includegraphics[width=\textwidth]{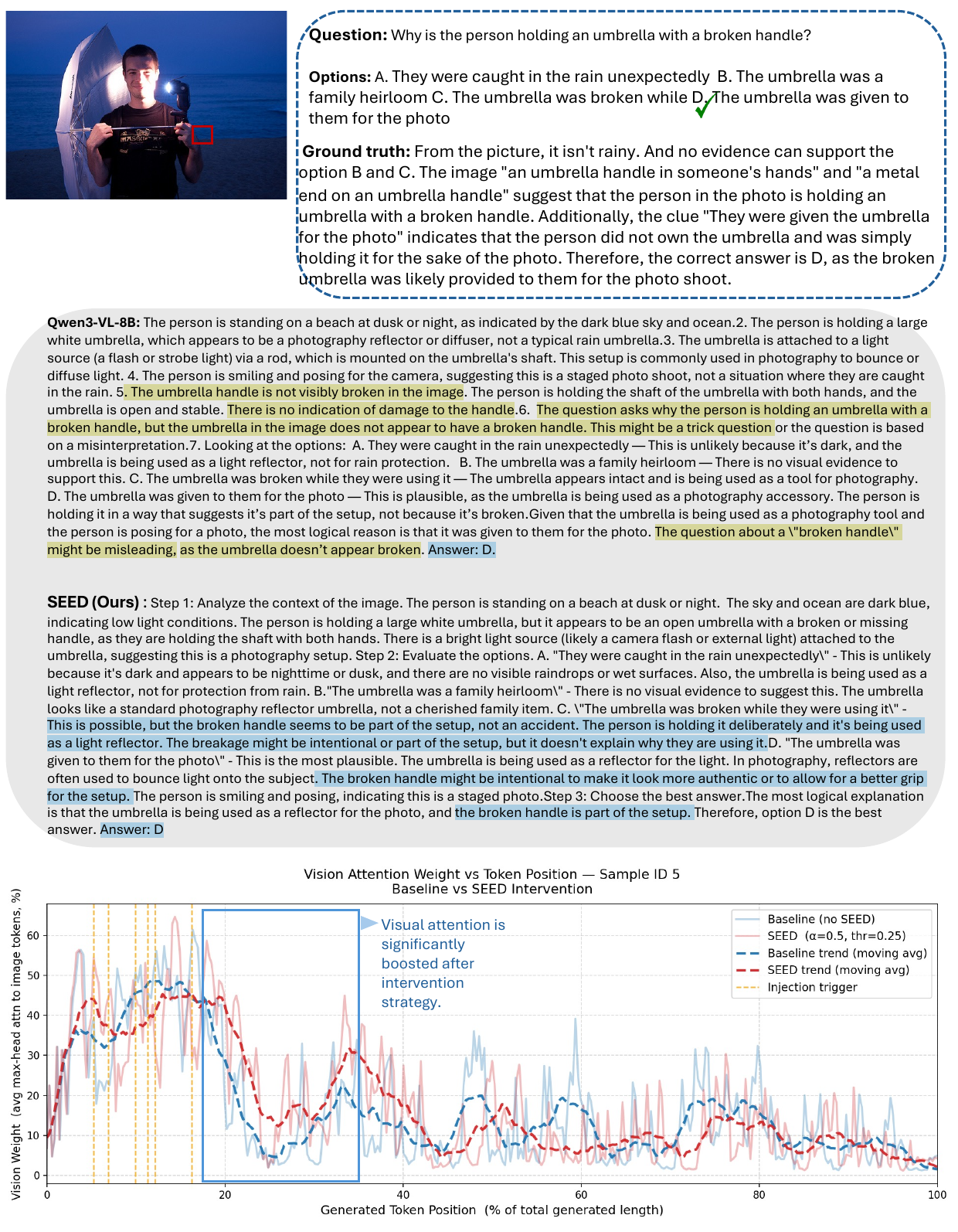}
    \caption{Full analysis on one example to demonstrate the effectiveness of SEED. The hallucination related text is highlighted in yellow and the statement related to correct reasoning is highlighted in blue.}
    \label{fig:more case}
\end{figure*}

Figure~\ref{fig:more case} provides a comprehensive case study of how SEED improves both reasoning quality and underlying modality interaction dynamics. From the top qualitative analysis, the baseline model produces partially unreliable reasoning, ignoring the fact that the \texttt{umbrella has a broken handle}, leading to hallucination-prone intermediate steps. In contrast, SEED generates more coherent and visually grounded reasoning, where key inference steps are supported by image evidence, resulting in the correct answer.

This behavioral difference is further reflected in the attention dynamics (bottom). As shown in the vision attention plot, the baseline exhibits relatively unstable attention with a slow decay, maintaining non-trivial attention weights even in later stages of generation. This suggests prolonged but ineffective reliance on visual information, which correlates with its inconsistent reasoning. In contrast, after the intervention trigger, SEED significantly boosts visual attention in the early-to-mid stages, followed by a sharper decay and stabilization, indicating more efficient integration of visual evidence into the reasoning process. Importantly, this pattern closely matches the attention dynamics identified in Figure~\ref{fig:preliminary1}(b), suggesting that the intervention effectively steers the model toward the desirable modality interaction regime.
\section{Limitations}\label{limitation}
Although SEED effectively mitigates hallucinations during multimodal reasoning, it has several limitations. First, the relationship between entropy dynamics and hallucination established in this work is correlational; SEED uses it as a detection signal without claiming causality. Second, the injected visual representation is query-independent, and aligning it with the cross-attention-based Llama-3.2-Vision backbone introduces an additional compatibility constraint associated with localized degradations (Appendix~\ref{app:failure}); architecture-specific feature alignment may further improve robustness. Third, while the fixed trigger policy simplifies deployment across settings, an adaptive policy may better capture model-specific reasoning dynamics. In particular, the model could learn to self-evaluate the trigger conditions and the attention reweighting strategy across different reasoning dynamics. Future work could investigate adaptive intervention mechanisms that learn to modulate visual re-alignment based on uncertainty evolution, reasoning trajectories, or model-specific behaviors.
\section{Social Impacts}\label{social impacts}
The social impacts of this work are primarily centered on improving the reliability and trustworthiness of multimodal AI systems deployed in real-world settings. By mitigating hallucinations in multimodal large language models, the proposed approach can enhance decision-support systems in high-stakes domains such as education, healthcare, and accessibility technologies, where incorrect or overconfident model outputs may lead to misleading conclusions or suboptimal decisions. Improved grounding interactions between visual and textual modalities also contribute to more transparent and interpretable AI behavior, which is critical for user trust and regulatory compliance. Furthermore, reducing hallucination-induced errors can lower the risk of misinformation propagation in AI-assisted content generation systems, thereby supporting healthier information ecosystems.

\end{document}